\documentclass[a4paper,11pt]{article}

\usepackage{bm}
\usepackage{color}
 \usepackage{amssymb}
 \usepackage{amsbsy}
 \usepackage{amscd}
 \usepackage{amsmath}
 \usepackage{amsthm}
\usepackage{amsmath}
\usepackage{amsfonts}

\newcommand{\RR}{\mathbb R}
\newcommand{\CC}{\mathbb C}
\newcommand{\EE}{\mathbb E}
\newcommand{\VV}{\mathbb V}

\begin{document}

\title{
Mathematical Theory of Bayesian Statistics
for Unknown Information Source}

\author{Sumio Watanabe
\\
Department of Mathematical and Computing Science
\\
Tokyo Institute of Technology 
\\
2-12-1 Oookayama, Meguro-ku,Tokyo  52-8552 Japan. 
mailbox W8-42\\
E-mail: swatanab@c.titech.ac.jp
}

\date{}

\maketitle

\begin{abstract}

In statistical inference, uncertainty is unknown and all models are wrong. 
 That is to say, a person who makes
a statistical model and a prior distribution is simultaneously aware that both are
fictional candidates. 
To study such cases, statistical measures have been
constructed, such as cross validation, information criteria, and 
marginal likelihood, however, 
their mathematical properties have not yet been completely clarified
when statistical models are under- and over- parametrized. 

We introduce a place of mathematical theory of Bayesian
statistics for unknown uncertainty, which clarifies 
general properties of cross validation, information criteria, 
and marginal likelihood, even if an unknown 
data-generating process is unrealizable by a model or even if
the posterior distribution cannot be approximated by any normal
distribution. Hence it gives a helpful standpoint for a person who
cannot believe in any specific model and prior. 

This paper consists of three parts.  The first is a new result, whereas
the second and third are well-known previous results 
with new experiments. 
We show there exists a more precise estimator of the generalization loss
than leave-one-out cross validation, there exists 
a more accurate approximation of marginal likelihood than BIC, 
and the optimal hyperparameters for generalization loss and marginal likelihood 
are different.

\end{abstract}

\pagebreak

\section{Introduction}

Bayesian inference is now widely employed in statistics and machine
learning, because it provides a precise  posterior  predictive distribution 
when statistical models and learning machines 
have hierarchical structures or hidden variables. 
In fact, the more complex statistical models are necessary in practical
applications, the more important Bayesian inference becomes. 
This is the reason why mathematical foundation of Bayesian statistics
is necessary for the cases 
when models and machines are under- or over-parametrized, 
in other words, they are too simple or too complex to minimize 
the generalization loss. 

In an older Bayesian statistics of the 20th century, it was said that 
a person should have ability to capture an uncertainty by a
statistical model and to represent a degree of belief by a prior distribution.
 Also it was said that, 
 if a model is made by synthesis of partial models, 
then the priors of partial models and their parameters should be determined. 
Nowadays, however, we know the set of all probability 
distributions is so large that a person cannot believe in any specific one
even for personal decision, 
which is often referred to as ``all models are wrong'' \cite{Box1976}. 
It was pointed out that a method how to determine or criticize a statistical model
still remains unsolved \cite{Smith}. 
In decision theory, coherent inference based on the subtle 
misspecification of a statistical model may take us to the wrong conclusion 
though it has been carefully prepared 
 \cite{Binmore2017}.  Hence, in Bayesian statistics, we need to 
check the posterior predictive distribution and improve 
 a model as sample size increases, in other words, 
a statistical model that is fixed before observation of a sample 
is not optimal in general \cite{Gelman2013}. 
In the use of Bayesian statistics
 for scientific purpose, preparation of a model and a prior 
 needs rethinking outside of a small world \cite{McElreath2020}. 
A new paradigm was proposed that both a statistical model and a prior distribution 
are candidate systems which had better be optimized for unknown data-generating
process by mathematical procedures
 \cite{Akaike1974,Akaike1980,Akaike1980a}. Based on these researches 
 in the computer age \cite{Efron}, it was proposed that modern statistics
is based on both computational algorithms and inferential evaluation, which  
has been accepted in statistics, data science, and machine learning fields. 

In this paper, we introduce a place of mathematical foundation of Bayesian 
statistics for the case when uncertainty is unknown 
in a large world.  That is to say, we study a case when 
 a person, who makes a pair of a statistical model and
a prior distribution, is simultaneously aware 
that it is  only a  fictional  candidate. 
Needless to say, there is no completely objective evaluation method, 
however, it is possible to prepare a much wider set of probability distributions 
which contains a person's choice as a special one and to examine
a candidate pair from a more generalized point of view. 

Statistical evaluation measures for unknown date-generating process have been 
proposed, for example, cross validation,
information criteria, and marginal likelihood.  
However, their mathematical properties are not yet fully clarified, 
because, if a statistical model has hierarchical structure or latent variables, 
the posterior distribution is highly singular. In this paper, 
we clarify general theory when a statistical model is 
under- or over- parametrized, and show the following three mathematical properties. 
 The first part is a new result, whereas
 the second and third parts are well-known previous results 
 with new experiments. 

First, although the leave-one-out cross validation \cite{Gelfand1992,Vehtari1,Vehtari2}
 and information criterion \cite{jmlr1}
have the same asymptotic expectation values as the generalization loss, 
it is clarified that
they have inverse correlation to the generalization loss. 
Hence neither leave-one-out cross 
validation nor information criterion is the best estimator of the generalization
loss in general. In this paper, we show the better estimator can be made by using an
 adjusted cross validation which is a weighted sum 
of the leave-one-out and  the hold-out cross validations or 
the out-of-sample validation whose concrete definition is given in
section \ref{section:LOOWAIC}, eq.(\ref{eq:Hn}). 

Second, we study the asymptotic behavior of the marginal likelihood 
when the posterior distribution is far from an normal distribution. 
If the posterior distribution can be approximated by some normal 
distribution, then the free energy, which is the minus log marginal
likelihood, can be asymptotically approximated by BIC \cite{Schwarz1978}
whose variance term is given by the half of dimension of the parameter
space.  We show that, if the posterior distribution contains singularities, 
the half of the dimension of the parameter space in BIC is replaced by 
the real log canonical threshold whose definition is introduced in 
section \ref{section:framework}, eq.(\ref{eq:zeta}). 
We also study several methods how to estimate
 the asymptotic free energy in such general cases. 

Third, we compare the leave-one-out cross
validation, the widely applicable information criterion, and the marginal 
likelihood as measures of a prior distribution. 
When a candidate prior distribution is a parametric function of a hyperparameter, 
the optimal hyperparameter that minimizes the leave-one-out 
cross validation or the widely applicable information criterion converges to the optimal 
one that minimizes the average generalization loss, whereas the optimal 
hyperparameter that maximizes the marginal likelihood does not. 
Their equivalence and difference are studied theoretically and experimentally. 

In section \ref{section:framework}, 
we explain a place of mathematical framework of the Bayesian
statistics when uncertainty is unknown. 
Two statistical measures, the free energy and the
generalization loss are defined when data-generating process is unknown. 
Also the definition and meaning of the real log canonical threshold (RLCT) are
introduced, which plays the central role in the theory of Bayesian statistics. 
In section \ref{section:LOOWAIC}, we explain the probabilistic properties of
the generalization loss, leave-one-out cross validation, and widely applicable information criterion, 
and propose the more precise measure of the generalization loss can be made. 
In section \ref{section:free}, the free energy or the minus log marginal likelihood is studied. 
In singular cases of overparametrized models, they cannot be
 approximated by BIC, but 
can by new algebraic geometrical studies. 
In section \ref{section:prior}, a prior optimization problem in 
a regular statistical model is
analyzed. In sections \ref{section:discuss} and \ref{section:conclusion},
the results of this paper are discussed and concluded.

\section{Mathematical framework of Bayesian 
statistics for unknown uncertainty}
\label{section:framework}

In this section, we introduce a mathematical framework 
of Bayesian statistics for a case when uncertainty is unknown. 

Let $n$ be an arbitrary positive integer and 
$x^n=\{x_i\in\RR^N\;;\;i=1,2,...,n\}$ be a set of real vectors.  
In this paper, we study a case when 
a statistical model is parametric and the set of parameters 
is a subset of a finite dimensional Euclidean space.  
We assume that a person, an agent, or an artificial intelligence
makes  a candidate pair of a statistical model 
$p(x|\theta)$ and a prior distribution $\pi(\theta)$, 
where $\theta\in\Theta\subset \RR^d$ is 
a parameter, such that 
\begin{align}
\theta&\sim\pi(\theta),\label{eq:1}
\\
x^n&\sim \prod_{i=1}^{n} p(x_i|\theta).\label{eq:2}
\end{align}
Then the posterior and  posterior 
predictive distributions based on eqs.(1) and (2)
are automatically defined by
\begin{align}
p(\theta|x^n)&=\frac{1}{p(x^n)}\pi(\theta)\prod_{i=1}^n p(x_i|\theta),
\label{eq:post}
\\
p(x|x^n)&=\int p(x|\theta)p(\theta|x^n)d\theta,
\label{eq:pred}
\end{align}
where $p(x^n)$ is the marginal likelihood,
\begin{align}\label{eq:p(xn)}
p(x^n)=\int \pi(\theta)\prod_{i=1}^np(x_i|\theta)d\theta.
\end{align}
We study a case when a person is aware 
that a candidate pair $p(x|\theta)$ and
$\pi(\theta)$ is a  fictional  one and wants to evaluate them
from a more general standpoint. 
It is impossible to evaluate it from the completely objective 
point of view, however, we show that a much more objective place can be set
than a specifically fixed pair. 
From eqs.(1) and (2), it follows that $x^n$ is exchangeable as shown by eq.(\ref{eq:p(xn)}). 
If $x^n$ $(n=1,2,3,...)$ is changeable, then by de Finetti's theorem, 
there exist an unknown probability distribution $q(x)$ and 
 an unknown functional probability distribution ${\cal Q}(q)$ such that
\begin{align}
q(x)&\sim {\cal Q}(q),\label{eq:3}
\\
x^n&\sim \prod_{i=1}^{n} q(x_i). \label{eq:4}
\end{align}
Then the pair in eqs.(\ref{eq:1}) and (\ref{eq:2}) made by a person 
is a special case of the general one in 
 eqs.(\ref{eq:3}) and (\ref{eq:4}). Hence,
if a person who made eqs.(\ref{eq:1}) and (\ref{eq:2})
rejects the existence of eqs.(\ref{eq:3}) and (\ref{eq:4}), it 
is a mathematical contradiction. 
In other words, if a person
prepares a statistical model $p(x|\theta)$ and a prior distribution $\pi(\theta)$, 
then $x^n$ is understood as a realization of  
a set of independent random variables $X^n$ whose 
probability distribution is unknown $q(x)$ 
which is subject to unknown ${\cal Q}(q)$. 
Hereafter, 
we use the capital notation $X^n$ for such a random variable,
and the pair $q(x)$ and $Q(q)$ is referred to as an unknown
information source or an unknown data-generating process.

\vskip3mm
\noindent{\bf Note.} A sample $x^n$ is understood to be from
 a real world, whereas a pair
$p(x|\theta)$ and $\pi(\theta)$ is a 
fictional  candidate prepared by a person.
Both $q(x)$ and ${\cal Q}(q)$ may be interpreted as also 
abstract concepts in  
a person's mathematical mind, or an scientific assumption of date-generating process
 in the real world. 
It should be emphasized that the same mathematical theory holds, 
independently of the assumption that $q(x)$ and ${\cal Q}(q)$ are real or unreal,
because mathematical framework is constructed for an arbitrary $p(x|\theta)$, 
$\pi(\theta)$, $q(x)$, and ${\cal Q}(q)$. 
In both cases, if a pair $p(x|\theta)$ and $\pi(\theta)$ is provided, then the
existence of $q(x)$ and ${\cal Q}(q)$ is automatically derived. 
A person who cannot accept the existence of $q(x)$ or ${\cal Q}(q)$
should reject the candidate pair $p(x|\theta)$ and $\pi(\theta)$.  
For interpretation of the case studied in this paper, see subsection 
\ref{subsection:meaning}. 
In this paper, we mainly study exchangeable cases. 
For cases when a sample is neither independent nor exchangeable, 
see subsection \ref{subsection:LOOdiscuss}.
\vskip3mm
For an arbitrary function $f(X^n)$ of $X^n$, 
its expectation value for  $q(x)$ is defined by 
\[
\EE[ f(X^n) |q]=\int  f(x^n) \prod_{i=1}^n q(x_i)dx_i. 
\]
The average and empirical entropies are respectively defined by
\begin{align}
S(q)&=-\int q(x)\log q(x)dx,\label{eq:S}
\\
S_n(q)&=-\frac{1}{n}\sum_{i=1}^n\log q(X_i).\label{eq:Sn}
\end{align}
Two well-known functionals of 
 $(p,\pi)\equiv (p(x|\theta),\pi(\theta))$, 
the free energy $F_n=F_n(p,\pi)$  and the generalization loss $G_n=G_n(p,\pi)$, 
which are the minus log marginal likelihood and the minus expected log likelihood
respectively,  are defined by
\begin{align}
F_n&=-\log \int \pi(\theta)\prod_{i=1}^n p(X_i|\theta) d\theta, \label{eq:Fn}
\\
G_n&=-\int q(x)\log p(x|X^n) dx. \label{eq:Gn}
\end{align} 
Then $\EE[F_n|q]$ satisfies
\begin{align}
\EE[F_n|q]={\rm KL}(q(x^n)||p(x^n))+nS(q),
\end{align}
where  ${\rm KL}(q(x^n)||p(x^n)) $ is the simultaneous Kullback-Leibler divergence
\[
{\rm KL}(q(x^n)||p(x^n))=\int q(x^n)\log\frac{q(x^n)}{ p(x^n)}dx^n,
\]
and $q(x^n)=\prod_{i=1}^n q(x_i)$. Therefore $\EE[F_n|q]$ is minimized at
$q(x^n)=p(x^n)$. Also 
$
\EE[ G_n |q]
$
satisfies
\begin{align}
\EE[G_n|q]={\rm KL}(q(x)||p(x|x^n))+S(q),
\end{align}
where ${\rm KL}(q(x)||p(x|x^n))$ is the conditional Kullback-Leibler divergence
\[
{\rm KL}(q(x)||p(x|x^n))=\int q(x)q(x^n)\log\frac{q(x)}{ p(x|x^n)}\;dx\;dx^n.
\]
Therefore $\EE[G_n|q]$ is minimized at
$q(x)=p(x|x^n)$. 
These two properties show that both the average free energy and 
the average generalization loss
can be employed as measures of appropriateness of the candidate pair given by 
eqs.(\ref{eq:1}) and (\ref{eq:2}) with respect to $q(x)$. 
In general, for an arbitrary positive integer $n$
\begin{align}
\EE[G_n|q]=\EE[F_{n+1}|q]-\EE[F_n|q]\label{eq:GnFn}
\end{align}
holds, however, minimization of the free energy gives the different result from that 
of the generalization loss.

Also we can define two functionals of a candidate pair
$(p(x|\theta),\pi(\theta))$, 
\begin{align}
{\cal F}(p,\pi)&=\int \EE[F_n|q] \;dQ(q),
\\
{\cal G}(p,\pi)&=\int \EE[G_n|q] \;dQ(q).
\end{align}
Then it follows that
\begin{align}
{\cal F}(p,\pi)&= - \int \overline{q}(x^n) \log \overline{q}(x^n) dx^n 
+\int \log \overline{q}(x^n) \log \frac{\overline{q}(x^n)}{p(x^n)}dx^n
\\
{\cal G}(p,\pi)&=
-\int \overline{q}(x^n)\overline{q}(x|x^n) \int \log \overline{q}(x|x^n)dx dx^n
\\
&
+\int \overline{q}(x^n)\overline{q}(x|x^n)\log \frac{ \overline{q}(x|x^n)}{p(x|x^n)}dx \;dx^n,
\end{align}
where 
\begin{align}
\overline{q}(x^n)&=\int q(x^n)dQ(q),
\\
\overline{q}(x|x^n)&=\frac{\int q(x)q(x^n)dQ(q)}{\int q(x^n)dQ(q)}.
\end{align}
Therefore
 both ${\cal F}(p,\pi)$ and ${\cal G}(p,\pi)$ 
are minimized if $q(x)=p(x|\theta)$ and $Q(q)=\pi(\theta)$. 
If a candidate pair $(p(x|\theta),\pi(\theta))$
 happens to be completely equal to the unknown $(q(x),Q(q))$, 
then it is optimal from both viewpoints of ${\cal F}(p,\pi)$ and
${\cal G}(p,\pi)$. However, if a candidate one is not completely equal to the 
unknown one, two evaluations ${\cal F}(p,\pi)$ and ${\cal G}(p,\pi)$ are not
equivalent to each other.

For given probability distributions $p(x|\theta)$ and $q(x)$, 
the average and empirical log loss functions, which are equal to
the average and empirical average of the minus log likelihood functions
respectively,
  are defined by 
\begin{align}
L(\theta)&=-\int q(x)\log p(x|\theta)dx,\label{eq:L(theta)}
\\
L_n(\theta)&=-\frac{1}{n}\sum_{i=1}^n\log p(X_i|\theta).\label{eq:Ln(theta)}
\end{align}
Let $\Theta_0$ be the set of all parameters that minimize $L(\theta)$, 
and $\theta_0$ be an element of $\Theta_0$. 
In general, $\Theta_0$ may consist of multiple elements and the
Hessian matrix of $L$ need not be positive definite. 
If $\Theta_0$ consists of a single element and the Hessian matrix
at $\theta_0$ is positive definite, then $q(x)$ is said to be regular for $p(x|\theta)$, 
if otherwise singular. 
For example, in overparametrized models such as normal mixtures, neural networks, 
and many learning machines, 
$\Theta_0$ is an analytic or algebraic set with singularities \cite{Watanabe2009}. 
In singular cases, the posterior distribution can not be approximated by
any normal distribution even if $n$ is sufficiently large. 
In this paper we assume that $\Theta_0$ consists of multiple elements in general
and that 
 $p(x|\theta_0)$ does not depend on the choice
of $\theta_0\in\Theta_0$.  Two variables $L_0$ and $L_{0,n}$ are defined by
$L_0=L(\theta_0)$ and $L_{0,n}=L_n(\theta_0)$, which are called the
average and empirical log losses of a parameter $\theta_0$, respectively. 
For the case when $p(x|\theta_0)$ 
depends on $\theta_0\in\Theta_0$, see \cite{Nagayasu,renormalizable}. 
If there exists $\theta_0$ such that 
$q(x)=p(x|\theta_0)$, then $q(x)$ is said to be realizable by $p(x|\theta)$.
If $q(x)$ is realizable by $p(x|\theta)$, then $S(q)=L_0$ and $S_n(q)=L_{0,n}$. 
The following concept RLCT is defined in both cases when
$q(x)$ is unrealizable by and singular for a statistical model $p(x|\theta)$. 

\vskip3mm\noindent
{\bf Definition of RLCT}. 
For a given triple, $(p(x|\theta),\pi(\theta),q(x))$, 
a zeta function of Bayesian statistics is defined by
\begin{align}\label{eq:zeta}
\zeta(z)=\int (L(\theta)-L_0)^z\pi(\theta)d\theta,
\end{align}
where $z\in\CC$ is one complex variable. If $L(\theta)$ is 
a piecewise analytic function of $\theta$, then $\zeta(z)$ is a
holomorphic function in the region $\Re(z)>0$, which can 
be analytically continued to the unique meromorphic function 
on the entire complex plane \cite{Atiyah,Hironaka,Watanabe2009}. 
It is proved that all poles of $\zeta(z)$ are real and negative 
numbers. Let $-\lambda$ $(\lambda>0)$ be the largest pole of $\zeta(z)$
and $m$ be its order. 
The constants $\lambda$ and $m$ are called the real log
canonical threshold (RLCT) and multiplicity. 
\vskip3mm
The concept RLCT is 
a well-known birational invariant in algebraic geometry, which
plays an important role also in Bayesian statistics. 
If $q(x)$ is regular for a statistical model, then
$\lambda=d/2$ and $m=1$, where $d$ is the dimension of the 
parameter space. For examples of RLCTs of singular statistical models, 
see subsection \ref{subsection:RLCT}. 
\vskip3mm\noindent
{\bf Geometric Understanding of RLCT.}  It has a clear geometric
meaning. In fact, we can prove \cite{Watanabe2009} that 
\[
\lambda=\lim_{\varepsilon\rightarrow +0}
\frac{\log{\rm Prob}(\varepsilon)}{\log\varepsilon},
\]
where ${\rm Prob}(\varepsilon)$ is the probability of
the set of almost optimal parameters measured by the 
prior distribution, 
\[
{\rm Prob}(\varepsilon)
=\int_{L(\theta)<L_0+\varepsilon}
d\pi(\theta).
\]
Since this probability is equal to an invariant of singularities, 
 the statistical estimation performance of the Bayesian
inference is determined by a kind of volume dimension of an analytic or an 
algebraic set. 
Note that, the smaller the probability is, the larger RLCT is,
since  $\log \varepsilon\rightarrow -\infty$. It was proved that, 
if $0<\pi(\theta)<\infty$ on $\Theta_0$, then $\lambda$ 
does not depend on the choice of $\pi(\theta)$ and 
singularities in $\Theta_0$ make the probability larger than the regular points, hence
RLCT smaller. If $\pi(\theta_0)=0$ or
 $ \infty$ at $\theta_0\in\Theta_0$ 
by controlling a hyperparameter,  
then $\lambda$ depends on the hyperparameter. 
It may be an important fact that Jeffreys' prior is equal to zero at singularities
because the Fisher information matrix contains zero eigenvalue.

\section{Cross Validation and Information Criterion}
\label{section:LOOWAIC}

\subsection{Adjusted Cross Validation}

In this section, we study statistical inference about the generalization loss $G_n$.
Let us define a training loss $T_n$, a leave-one-out cross validation loss (LOO)
$C_n$, and a widely applicable information criterion (WAIC) 
$W_n$ respectively by
\begin{align}
T_n&=-\frac{1}{n}\sum_{i=1}^n \log p(X_i|X^n),\label{eq:Tn}
\\
C_n&=-\frac{1}{n}\sum_{i=1}^n \log p(X_i|X^n\setminus X_i),\label{eq:Cn}
\\
W_n&=T_n
+\frac{1}{n}\sum_{i=1}^n \VV_\theta[\log p(X_i|\theta)],\label{eq:Wn}
\end{align}
where $p(x|X^n)$ is the  posterior  predictive distribution defined by eq.(\ref{eq:pred}), 
$X^n\setminus X_i$ is the sample 
leaving $X_i$ out from $X^n$, and 
$\VV_\theta[\;\;]$ is the variance about $\theta$ in the posterior distribution
eq.(\ref{eq:post}),  which  is, for an arbitrary function $f(\theta)$,
\[
\VV_\theta[f(\theta)]=
\int f(\theta)^2 p(\theta|X^n) d\theta
- \left(\int f(\theta) p(\theta|X^n) d\theta\right)^2.
\] 
Note that $T_n$, $C_n$, and $W_n$ are defined without 
using any information about $q(x)$. 

Then even if $q(x)$ is not realizable by $p(x|\theta)$ and even if $q(x)$ is
singular for $p(x|\theta)$, the following properties
are proved \cite{jmlr1,wata2018}. Based on algebraic geometrical method, 
it was proved that 
there exist random variables $R_1$ and $R_2$ such that
\begin{align}
G_n&=L_0+\left( \lambda+R_1-R_2\right)\frac{1}{n}+o_p(1/n),\label{eq:GnTh}
\\
T_n&=L_{0,n}+\left( \lambda-R_1-R_2\right)\frac{1}{n}+o_p(1/n),\label{eq:TnTh}
\\
C_n&=L_{0,n}+\left( \lambda-R_1+R_2\right)\frac{1}{n}+o_p(1/n),\label{eq:CnTh}
\\
W_n&=L_{0,n}+\left( \lambda-R_1+R_2\right)\frac{1}{n}+o_p(1/n),\label{eq:WnTh}
\end{align}
where $R_1$ and $R_2$ satisfy $\EE[R_1]=\EE[R_2]+o(1)$ and 
\[
R_2=\frac{1}{2}\sum_{i=1}^n\VV_\theta[\log p(X_i|\theta)]+o_p(1). 
\]  
The concrete forms of $R_1$ and $R_2$ are given in \cite{wata2018},
in Theorem 6 for regular cases and Theorem 14 for singular cases. The 
convergences in distribution of $R_1$ and $R_2$ is proved using 
the empirical process theory using  renormalized log likelihood functions. 
Both $C_n$ and $W_n$ are asymptotically unbiased estimators
of $G_n$, whereas neither  AIC  nor DIC \cite{Spiegel2002} is.
Note that, if $X^n$ is a set of independent random variables, then LOO and WAIC
are asymptotically equivalent, if otherwise not.  
For the differences between LOO and WAIC, see subsection
 \ref{section:discuss}.

These equations (\ref{eq:GnTh}) - (\ref{eq:WnTh}) 
show a good behavior of 
the leave-one-out cross validation and the widely applicable information criterion, 
 however, $G_n-L_0$ and $C_n-L_{0,n}$ have an inverse correlation 
to each other \cite{jmlr1,wata2018},
\[
\left(G_n-L_0\right)+\left(C_n-L_{0,n}\right)=\frac{2\lambda}{n}
+o_p\left(\frac{1}{n}\right). 
\]
Also $G_n-L_0$ and $W_n-L_{0,n}$ satisfy the same equation, which
is the disadvantage of both LOO and WAIC. 
It should be emphasized that many statisticians may not be aware this weak 
point because it is not trivial from the definitions of LOO and WAIC. 
The inverse correlations of $C_n-L_{0,n}$ and $W_n-L_{0,n}$ with $G_n-L_0$ 
are asymptotic properties whose proofs are given theoretically and 
intuitive explanation is not yet given. Although both LOO and WAIC are the good estimators
of the generalization loss in practical applications, 
they are not always the best estimators in general and might be improved more.  
In this section we mainly study how to improve this disadvantage. 

In order to overcome the disadvantage of LOO and WAIC, a more precise estimator
 can be made based on the 
theoretical results, eqs. (\ref{eq:GnTh})-(\ref{eq:WnTh}).
For given two positive integers $n_1$ and $n_2$ which satisfies $n=n_1+n_2$, 
we divide a sample $X^n=(X^{n_1},X^{n_2})$, where
$X^{n_1}=(X_1,X_2,...,X_{n_1})$ and
$X^{n_2}=(X_{n_1+1},X_{n_1+2},...,X_n)$.
Let $L_{0,n_1}$ and $L_{0,n_2}$ be the log loss of a parameter $\theta_0$ 
defined in eq.(\ref{eq:Ln(theta)}) using $X^{n_1}$ and $X^{n_2}$ instead of $X^n$ respectively. 
The leave-one-out cross validation using $X^{n_1}$ is 
\[
C_{n_1}=-\frac{1}{n_1}\sum_{i=1}^{n_1}\log p(X_i|X^{n_1}\setminus X_i).
\]
Then it follows that
\[
\EE[C_{n_1}]=L_0+\frac{\lambda}{n_1}+o(1/n_1).
\]
The hold-out cross validation or the out-of-sample cross validation is
defined by
\begin{align}
H_{n_2}&=-\frac{1}{n_2}\sum_{i=n_1+1}^{n}
\log p(X_i|X^{n_1}), \label{eq:Hn}
\end{align}
which estimates the generalization loss of the  posterior 
predictive distribution
made by $X^{n_1}$ that is measured by using $X^{n_2}$. This is an 
unbiased estimator of $G_{n_1}$, 
\[
\EE[H_{n_2}]=L_0+\frac{\lambda}{n_1}+o_p(1/n_1). 
\]
An adjusted cross validation (ACV) is proposed by a weighted 
sum of the leave-one-out and hold-out cross validations, 
\begin{align}
A_{n}& =\frac{n_1}{n}C_{n_1}+\frac{n_2}{n}H_{n_2}.\label{eq:An}
\end{align}
By the definition, $A_{n}$ is an 
unbiased estimator of the generalization
loss $G_{n_1}$,
\[
\EE[A_{n}]=L_0 + \frac{\lambda}{n_1}+o(1/n_1).
\]
It follows that 
\[
\frac{n_1}{n} \EE[A_{n}-L_{0,n}]=\EE[G_n-L_0]+o(1/n),
\]
which shows that $(n_1/n)(A_n-L_{0,n})$ is an asymptotic unbiased estimator 
of $G_n-L_0$. Moreover, $A_n-L_{0,n}$ does not have an inverse correlation
to $G_{n_1}-L_0$, hence it provides the better estimator than 
$C_n-L_{0,n}$ and $W_n-L_{0,n}$, 
if the average of the $1/n$ order term of $G_{n_1}$ is equal to that of $G_{n}$. 
 
For theoretical foundation in regular cases see subsection \ref{subsection:3.3}.
Note that, 
in the evaluation  of prior distributions on the condition that
 a statistical model is fixed, 
then the $1/n$ order terms are common and the compared 
terms have higher order than $1/n$, 
resulting that the adjusted cross validation 
may not be a better measure than LOO and WAIC. 
The higher order probabilistic behaviors of the generalization loss, LOO, and 
WAIC are shown in 
 section \ref{section:prior}. 

The rescaled hold-out or out-of-sample cross validation error
\begin{align}
\frac{n_1}{n}(H_{n_2}-L_{0,n_2})&=-\frac{1}{n}\sum_{i=n_1+1}^{n}
\left\{
\log p(X_i|X^{n_1})- \log q(X_i)\right\}
\nonumber
\end{align}
is also an unbiased estimator of $G_{n}-L$, which has the larger variance  
than $(n_1/n)(A_n-L_n)$. 
Note that, if a probability distribution $q(x)$ is realizable by $p(x|\theta)$,
then neither $L_0=S(q)$ nor $L_{0,n}=S_n(q)$ depends on $p(x|\theta)$ and $\pi(\theta)$. 

\subsection{Numerical Experiment}

In this subsection, we show a numerical experiment. 

\noindent
{\bf Example.1.} In order to illustrate the differences of the several estimators
of the generalization loss, a matrix factorization problem is studied. 
Let $X$, $A$, and $B$  be $M\times N$, $M\times H$, and $H\times N$
matrices respectively.  A statistical model and a prior distribution of a matrix
factorization are defined by
\begin{align}
p(X|A,B)&\propto\exp\left(-\frac{1}{2}\|X-AB\|^2\right),
\\
\pi(A,B)&\propto\exp\left(-\frac{1}{2\rho^2}\|A\|^2-\frac{1}{2\mu^2}\|B\|^2
\right),
\end{align}
where $\|\;\;\|$ is the Frobenius norm and $\rho,\mu>0$ are
hyperparameters. This model is sometimes employed for the purpose
that a random matrix $X$ is estimated by a product of low rank matrices 
$A$ and $B$. Since the map $(A,B)\mapsto p(x|A,B)$ is not one-to-one,
this model is not regular but singular. The posterior distribution cannot be
estimated by any normal distribution. 
The real log canonical threshold $\lambda$ of this model 
is same as that of the reduced rank regression which was clarified in 
\cite{Aoyagi2005}. In the experiment, we studied a case 
when $M=N=8$, $\rho=\mu=10$. 
A sample $X^n$ $(n=200)$ was taken from $q(X)=p(X|A_0,B_0)$ where 
$A_0B_0=\mbox{diag}(1,1,0,0,0,0,0,0)$ is a diagonal matrix. Hence
the rank of $A_0B_0$ is $H_0=2$. We conducted two experiments $H=2$ and $H=6$. 
If $H=2$, the model
is appropriate for $H_0$, but if $H=6$, the model is  over-parametrized. 
The values RLCTs for $H=2$ and $H=6$ are $\lambda=14$ and
$\lambda=24$ respectively, whereas half of the dimensions of the parameter spaces 
$d/2=(MH+HN-N^2)/2$ are $14$ and $30$, respectively. Two hundred independent
trials were conducted. 
 In experimental comparison, the average and 
empirical entropies of $q(X)$ 
 are reduced from the estimators. Let
the generalization error be ${\rm GE. E.}=G_n-S$, 
the cross validation error, the widely applicable 
information criterion error, the adjusted cross validation error,
the hold-out cross validation error, the AIC error, and the DIC \cite{Spiegel2002} error
are respectively defined by reducing the empirical entropy. 
\begin{align}
{\rm  LOO. E.}&=C_n-S_n(q), 
\\
{\rm WAIC. E.}&=W_n-S_n(q), 
\\
{\rm AC. E.}&=(n_1/n)(A_n-S_n(q)), 
\\
{\rm HO. E.}&=(n_1/n)(H_{n_2}-S_{n_2}(q)),
\\
{\rm AIC. E.}&={\rm AIC}-S_n(q),
\\
{\rm DIC. E.}&={\rm DIC}-S_n(q).
\end{align}
Note that $A_n-S_n(q)$ and $H_{n_2}-S_{n_2}(q)$ are estimators of $\EE[G_{n_1}]-S(q)$,
they are used as estimators of  $\EE[G_{n}]-S(q)$ by rescaling. In calculation of
the adjusted and hold-out cross validations, we used $n_1=n_2=100$. 
In Table.\ref{table:111},   in the case $H=2$, 
the six values from GE. E. to AIC. E. have asymptotically the same expectation 
values $\lambda/n=0.07$.   In the case $H=6$, 
the five values from GE. E. to HO. E. have asymptotically the same expectation 
values
$\lambda/n=0.12$, whereas neither AIC nor DIC does. These results are caused 
by the fact that matrix
factorization is a singular statistical model and the posterior distribution 
can not approximated by any normal distribution. 
In the table, MEAN and STD show averages and standard deviations
of these random variables, and  RSE shows the
root square errors of the generalization error and estimated errors,
$\EE[({\rm GE. E.}-{\rm LOO. E.})^2]^{1/2}$,
$\EE[({\rm GE. E.}-{\rm WAIC. E.})^2]^{1/2}$, and so on. 
These experimental results show the adjusted cross validation
is a better estimator of the generalization loss than other estimators, 
because it has the smallest standard deviation and root square error
in both cases $H=2$ and $H=6$. 

\begin{table}[tb]
\begin{center}
\begin{tabular}{|c|c|c|c|c|c|c|}
\hline
\multicolumn{1}{|c|}{} &
\multicolumn{3}{|c|}{$H_0=H=2$} 
& \multicolumn{3}{c|}{$H_0=2$, $H=6$}\\
\cline{2-7}
      & Mean & Std  & RSE & Mean & Std  & RSE\\
\hline

 GE. E. &  0.071& 0.019 & &  0.117 &  0.025 & \\
\hline
 LOO. E. &  0.069 &  0.018&  0.037&  0.118 
&  0.024 & 0.049 \\
\hline
WAIC. E.& 0.069&  0.018& 0.037& 0.118 &  0.024& 
 0.049 \\
\hline
AC. E. &  0.071&  0.014& 0.031& 0.117 &  0.017&  0.039\\
\hline
HO. E. &  0.074 &  0.032&  0.043&  0.119 &  0.040& 
 0.053 \\
\hline
AIC. E. & 0.068& 0.018&  0.037& 0.150 &  0.028 &
 0.062 \\
\hline
DIC. E. & -5.492& 11.432& 12.688& -1.262 &
 0.505 &  1.469 \\
\hline
\end{tabular}
\end{center}
\caption{Comparison of Estimators of Generalization Error
\newline
GE.E.: Generalization error, \newline
LOO.E.: Leave-one-out error, \newline
WAIC.E.: Widely applicable information criterion error,\newline
AC.E.: Adjusted cross validation error, \newline
HO.E.: Hold-out cross validation error,\newline
AIC.E.: Akaike information criterion error, \newline
DIC.E.: Deviance information criterion error. 
}
\label{table:111}
\end{table}

\subsection{Variances of Cross Validations in Regular Case}\label{subsection:3.3}

In this subsection, we theoretically compare the adjusted cross validation $A_n$ and the
leave-one-out cross validation $C_n$ when $q(x)$ is regular for $p(x|\theta)$. 
Since  $(n_1/n)(A_n-L_{0,n})$ and $C_n-L_{0,n}$ 
are asymptotic unbiased estimators of the generalization error $G_n-L_0$, we
show that the 
variance of $(n_1/n)(A_n-L_{0,n})$ is smaller than that of 
$C_n-L_{0,n}$ on the assumption both $n_1/n$ and
$n_2/n$ converge to constants. For the case when 
$q(x)$ is not regular for $p(x|\theta)$, theoretical comparison is still an open problem. 

As we assumed the regularity condition, $\theta_0$ is unique and 
the Hessian matrix
$J=\nabla^2 L(\theta_0)$ is positive definite. We define a matrix $I$ by
\begin{align}\label{eq:matI}
I &=\int \nabla \log p(x|\theta_0)(\nabla \log p(x|\theta_0))^T q(x) dx, 
\end{align}
where $T$ shows the transpose of the vector. 
Then based on just the same method as chapter 4, Theorem 6, in \cite{wata2018},
the leave-one-out and hold-out cross validations, $C_{n_1}$ and $H_{n_2}$,
 are asymptotically derived as
\begin{align}
C_{n_1}&=L_{0,n_1}+\frac{1}{2n_1}\left\{-\|\xi_{n_1}\|^2+d+\mbox{tr}(IJ^{-1})
\right\}+O_p(n_1^{-3/2})
\\
H_{n_2}&=L_{0,n_2}
+\frac{1}{2n_1}\left\{-2\sqrt{\frac{n_1}{n_2}}\xi_{n_1}
\cdot\eta_{n_2}+\|\xi_{n_1}\|^2+d-\mbox{tr}(IJ^{-1})
\right\}
+O_p(n_1^{-3/2})
\end{align}
where $\xi_{n_1}$ and $\eta_{n_2}$ are $\RR^d$-valued independent
random variables, 
\begin{align}
\xi_{n_1} & \equiv
-J^{-1/2}\frac{1}{\sqrt{n_1}}\sum_{i=1}^{n_1} \nabla \log p(X_i|\theta_0),
\\
\eta_{n_2}&
\equiv
-J^{-1/2}\frac{1}{\sqrt{n_2}}
\sum_{i=n_1+1}^n \nabla \log p(X_i|\theta_0),
\end{align}
which converge  in distribution to the same
normal distribution with average $0$ and covariance matrix 
$J^{-1/2}IJ^{-1/2}$. Consequently
\begin{align}
\EE[\|\xi_{n_1}\|^2]&\rightarrow \mbox{tr}(IJ^{-1}) ,\label{eq:xin2}
\\
\EE[\|\xi_{n_1}\|^4]-\EE[\|\xi_{n_1}\|^2]^2&\rightarrow 
2\mbox{tr}((IJ^{-1})^2),\label{eq:xin4}
\\
\EE[(\xi_{n_1}\cdot\eta_{n_2})^2]
& \rightarrow 
\mbox{tr}((IJ^{-1})^2).\label{eq:xieta}
\end{align}
By the definition of $C_n$ in eq.(\ref{eq:Cn}) and $A_n$ in eq.(\ref{eq:An}),
the asymptotic variances of $C_n-L_{0,n}$ and $(n_1/n)(A_n-L_{0,n})$ are 
given by eqs.(\ref{eq:xin2})-(\ref{eq:xieta}), resulting that 
\begin{align}
\VV[C_n-L_{0,n}] & =\frac{1}{2n^2}\mbox{tr}((IJ^{-1})^2)+o(n^{-2}),
\\
\VV[(n_1/n)(A_n-L_{0,n})]
&=\frac{1}{2n^4}
(n_1^2+n_2^2)\mbox{tr}((IJ^{-1})^2)+o(n^{-2}).
\end{align}
Hence the asymptotic variance of $(n_1/n)(A_n-L_{0,n})$ is made
smallest when $n_1=n_2=n/2$ and the variance in the smallest case is half of that of 
$C_n$.

\vskip3mm\noindent{\bf Remark.} As a consequence of this theoretical
result, if $q(x)$ is regular for $p(x|\theta)$, the same method can be 
extended. 
Let $n$ be quite large and let $A_{n/2}$ and $A'_{n/2}$ be 
adjusted cross validations calculated from $(X_1,X_2,...,X_{n/2})$ and
$(X_{n/2+1},X_{n/2+2},...,X_n)$ respectively. Then the following 
three random variables have asymptotically the same expectation value as
$\EE[G_n]-L_0\approx \mbox{tr}(IJ^{-1})/(2n)$, 
\begin{align}
& C_n-L_{0,n},
\\
& (1/2)(A_n-L_{0,n}),
\\
& (1/8)(A_{n/2}+A'_{n/2}-L_{0,n}),
\end{align}
whereas the ratio of their variances is $(1:1/2:1/4)$. The same method can be
generalized for arbitrary partition of a sample, resulting that the variance 
can be made arbitrary smaller. However, if 
a partition makes each sample size of each adjusted cross validation 
too small, the asymptotic evaluation 
 $\EE[G_{m}]-L_0\approx \mbox{tr}(IJ^{-1})/(2m)$ for
large $m$ may not be satisfied. It is the future study to clarify
the optimal number of partitions for a large $n$.

\section{Marginal Likelihood and Free Energy}\label{section:free}

In this section, several approximations of the free energy which 
 is the minus log marginal likelihood eq({\ref{eq:Fn}) are compared. 
Even if $q(x)$ is not realizable by $p(x|\theta)$ or even if
the posterior distribution cannot be approximated by
any normal distribution, it is derived \cite{Watanabe2009} that
\[
F_n=n L_{0,n}+\lambda\log n -(m-1)\log\log n+O_p(1). 
\]
where $\lambda$ and $m$ are RLCT and its multiplicity, respectively. 

By eq.(\ref{eq:GnFn}), the constant term of the free energy 
does not affect the generalization loss. In fact,
the asymptotic balances of biases and variances are given by
\begin{align}
\EE[F_n|q]&=n L_0+\lambda\log n,
\\
n\times \EE[G_n|q]&=nL_0+\lambda,
\end{align}
hence the variance term $\lambda\log n$ of the free energy
is larger than $\lambda$ of the generalization loss. In other words, 
optimizations for the free energy and the generalization loss are
incompatible with each other. According to the increase of 
sample size $n$, this the different balance of the bias and variance
affect the generalization performance \cite{wata2001}.

In general, it needs heavy computational cost to calculate the numerical value of the
free energy, hence several methods have been developed. 
If a statistical model is regular, 
then BIC \cite{Schwarz1978} is defined by
\[
{\rm BIC}=nL_{n}(\hat{\theta})+\frac{d}{2}\log n,
\]
where $\hat{\theta}$ is the maximum likelihood estimator. 
The asymptotically main term of the free energy ( AFE ) is given by
\[
{\rm  AFE }=nL_n(\hat{\theta})+\lambda\log n.
\]
In general, RLCT depends on $(q(x),p(x|\theta),\pi(\theta))$, this 
equation cannot be used directly if $q(x)$ is unknown. 
The generalized version of BIC onto singular cases was proposed by
using the estimated RLCT $\hat{\lambda}$ \cite{Drton}, 
\[
{\rm 
sBIC}=nL_n(\hat{\theta})+\hat{\lambda}\log n.
\]
This method sBIC needs theoretical results about RLCTs but 
does not need Markov chain Monte Carlo (MCMC) approximation of
the posterior distribution. 
By using the MCMC sample of the parameter, another method was 
developed. 
A posterior distribution of the inverse temperature $\beta>0$ is
defined by
\[
\EE_\theta^{(\beta)}[f(\theta)]=\frac{
\displaystyle
\int 
f(\theta)\pi(\theta)\prod_{i=1}^np(X_i|\theta)^{\beta}d\theta
}{\displaystyle
\int
\pi(\theta)\prod_{i=1}^np(X_i|\theta)^{\beta}d\theta
}.
\]
Then WBIC is defined by $\beta=1/\log n$ \cite{jmlr2,watanabe2021a}
\[
{\rm WBIC}=\EE_\theta^{(1/\log n)}[nL_n(\theta)],
\]
which satisfies 
\[
{\rm WBIC}=nL_{0,n}+\lambda\log n +o_p(\log n).
\]
Also it is proved that, if the posterior distribution
can be approximated by a normal distribution, then
\[
{\rm WBIC}={\rm BIC}+o_p(1). 
\]
Note that BIC, asymptotic free energy (AFE), and WBIC approximate 
the $\log n$ order term of the free energy, hence
their values have constant order differences from the free energy. 
The prior distribution does not affect their values directly if it does not change 
RLCT, whereas it does the constant order term. 
\vskip3mm\noindent
{\bf Example.2.} Let us compare BIC, WBIC, and 
asymptotic form of the free energy. The asymptotic form is
equal to sBIC if $\hat{\lambda}=\lambda$. 
A matrix factorization problem same as 
Example.1 was studied. For $M$, $N$, $H$, $H_0$, $n$ and $A_0B_0$, the 
same condition as Example.1 was conducted.  
We compared ${\rm AFE}/n-S_n(q)$, ${\rm BIC}/n-S_n(q)$, and 
${\rm WBIC}/n-S_n(q)$.
Table \ref{table:222} shows their averages and standard deviations. 
When $H=H_0$, then three values coincided with each others, whereas in a
singular case $H> H_0$, 
then they were different. The experimental result shows that WBIC  approximated 
the asymptotic from better than BIC in overparametrized cases. 

\color{red}

\begin{table}[tb]
\begin{center}
\begin{tabular}{|c|c|c|c|c|}
\hline
\multicolumn{1}{|c|}{} &
\multicolumn{2}{|c|}{$H_0=H=2$} 
& \multicolumn{2}{c|}{$H_0=2$, $H=6$}\\
\cline{2-5}
      & Mean & Std & Mean & Std  \\
\hline
 AFE$/n$ - $S_n$ &  0.30 &  0.02 &  0.48 &  0.03
\\
\hline
BIC$/n$ - $S_n$ &  0.30 &  0.02 &  0.64 &  0.03   \\
\hline
WBIC$/n$ - $S_n$ &  0.31 &   0.02 &  0.49 &   0.02 \\
\hline
\end{tabular}
\end{center}
\caption{Estimators of Free Energy\newline
 AFE$/n$ : Asymptotic free energy divided by $n$,\newline
BIC$/n$ : Bayesian information criterion divided by $n$, \newline
WBIC$/n$ : Widely applicable information criterion divided by $n$
}
\label{table:222}
\end{table}

\color{black}

\section{Evaluation of Prior Distributions}\label{section:prior}

In this section, we study the relation between prior distributions and 
generalization losses, on the assumption that a statistical 
model $p(x|\theta)$ is regular and fixed, that is to say, for an arbitrary $\theta$,
the Hessian matrix of $L(\theta)$ is positive definite.

We consider a condition that
a candidate prior distribution $\pi(\theta)\geq 0 $ may be improper, that is to say, 
in general, 
\[
\int \pi(\theta)d\theta=\infty.
\]
Even if it is improper, 
the posterior and  posterior  predictive distributions can be
defined by eqs. (\ref{eq:post}) and (\ref{eq:pred}) if both are finite. 
The generalization loss $G_n(\pi)$, the leave-one-out cross validation $C_n(\pi)$,  
the widely applicable information criterion $W_n(\pi)$, the 
hold-out cross validation $H_n(\pi)$ , and the adjusted cross validation $A_n(\pi)$ 
can also be defined by
the same eqs. (\ref{eq:Gn}), (\ref{eq:Cn}), (\ref{eq:Wn}). (\ref{eq:Hn}), and (\ref{eq:An}), respectively. 
However, if a prior distribution is improper, 
the marginal likelihood cannot be a measure of the appropriateness
of the prior distribution, because it may be made infinite by choosing an improper prior distribution. 

In this section we fix a statistical model $p(x|\theta)$ and 
study the effect of a candidate prior distribution $\pi(\theta)$.
The function $q(x)$ is assumed to be unrealizable by a statistical model in general. 
Let $\pi_0(\theta)>0$ be an arbitrary
fixed nonnegative function on $\RR^d$. For example, one can choose 
$\pi_0(\theta)\equiv 1$ for an arbitrary $\theta$. For a given 
$(\pi(\theta),\pi_0(\theta))$, a function $\phi(\theta)$ is defined by
\[
\phi(\theta)=\pi(\theta)/\pi_0(\theta).
\]
On the foregoing assumptions, 
it is proved in \cite{wata2018,watanabe2018}
 that there exists a function ${\cal M}(\phi,\theta)$ such that
\begin{eqnarray}
\EE[G_n(\pi)|q]&=&\EE[G_n(\pi_0)|q]+\frac{{\cal M}(\phi,\theta_0)}{n^2}
+o\Bigl(\frac{1}{n^2}\Bigr),\\
\EE[C_n(\pi)|q]&=&\EE[G_n(\pi_0)|q]+\frac{d/2+{\cal M}(\phi,\theta_0)}{n^2}+o\Bigl(\frac{1}{n^2}\Bigr),\\
\EE[W_n(\pi)|q]&=&\EE[G_n(\pi_0)|q]+\frac{d/2+{\cal M}(\phi,\theta_0)}{n^2}+o\Bigl(\frac{1}{n^2}\Bigr),
\end{eqnarray}
where $\theta_0$ is the parameter that minimizes $L(\theta)$ in eq.(\ref{eq:L(theta)}). 
That it to say, the expectation values of the generalization loss, the leave-one-out cross 
validation, and WAIC are 
equivalent in the higher order. 
Also it was proved in \cite{wata2018,watanabe2018}
that there exists a function $M(\phi,\theta)$ of $\phi(\theta)$ and $\theta$ which satisfies
\begin{eqnarray}
C_n(\pi)&=&C_n(\pi_0)+\frac{M(\phi,\hat{\theta})}{n^2}+o_p(\frac{1}{n^2}),\label{eq:LOO2}\\
W_n(\pi)&=&W_n(\pi_0)+\frac{M(\phi,\hat{\theta})}{n^2}+o_p\Bigl(\frac{1}{n^2}\Bigr), \label{eq:WAIC2}
\end{eqnarray}
where $\hat{\theta}$ is the parameter that 
maximizes $\pi_0(\theta)\prod_{i=1}^n p(X_i|\theta)$, 
in other words, $\hat{\theta}$ is 
 the {\it maximum a posteriori} estimator for a fixed prior function
 $\pi_0(\theta)$. 
Hence LOO and WAIC are also equivalent in the higher order as random variables. 
The concrete forms of the functionals ${\cal M}(\phi,\theta)$ and 
$M(\phi,\theta)$ are defined by using
higher order differential geometric forms  of the log density function 
$\log p(x|\theta)$
 \cite{wata2018,watanabe2018}. 
They satisfy 
\begin{eqnarray}
M(\phi,\hat{\theta})&=&{\cal M}(\phi,\theta_0)
+O_p\Bigl(\frac{1}{n^{1/2}}\Bigr), \label{eq:MM}\\
M(\phi,\EE_{\theta}[\theta])&=&M(\phi,\hat{\theta})
+O_p\Bigl(\frac{1}{n}\Bigr), \label{eq:MM2}\\
\EE[M(\phi,\hat{\theta})]&=&{\cal M}(\phi,\theta_0)
+O\Bigl(\frac{1}{n}\Bigr).\label{eq:EMM}
\end{eqnarray}
It should be emphasized that 
the generalization loss as a random variable has the different behavior from
its own average \cite{watanabe2018}
\begin{align}
G_n(\pi)
&= G_n(\pi_0)+O_p\Bigl(\frac{1}{n^{3/2}}\Bigr).\label{eq:G} 
\end{align}
The parameter that minimizes the average generalization loss does not 
minimize the random generalization loss. 
Also minimization of WAIC or LOO by choosing $\pi(\theta)$ 
makes the average generalization loss $\EE[G_n(\pi)]$ minimized 
asymptotically, however, it does not minimize
the generalization loss $G_n(\pi)$ as a random variable.

The free energy or the marginal likelihood has the different behavior than
the cross validation and information criterion. 
Let $F_n(\pi)$ be the free energy of a proper
prior distribution. Assume that $\pi(\theta)$ and $\pi_0(\theta)$ 
satisfy  $\int \pi(\theta)d\theta=\int\pi_0(\theta)d\theta=1$. 
Then it follows that \cite{Watanabe2022}
\begin{align}
F_n(\pi) & =F_n(\pi_0)-\log\frac{\pi(\theta_0)}{\pi_0(\theta_0)}+o_p(1).
\\
& =F_n(\pi_0)-\log\frac{\pi(\hat{\theta}_{mle})}{\pi_0(\hat{\theta}_{mle})}+o_p(1),
\end{align}
where $\theta_{mle}$ is the maximum likelihood estimator. 
Hence the minimization of $F_n(\pi)$ is asymptotically equivalent
to maximization of $\pi(\hat{\theta}_{mle})$ at the maximum likelihood estimator. 

If a set of  candidate prior distributions is given by $\{\pi(\theta|a)\;;\;
a\}$, where $a$ is a hyperparameter, the optimal hyperparameter that is determined 
by minimization of $F_n(\pi(\theta|a))$ does not converges to the parameter
that minimizes the average generalization loss in general even if the sample size 
tends to infinity.  For example,  a statistical model $p(x|m,s)$ and a prior 
distribution $\pi(m,s|a)$, where $a>0$ is a hyperparameter, are given 
\begin{align}
p(x|m,s)&=\sqrt{\frac{s}{2}}\exp\left(-\frac{s}{2}(x-m)^2\right),
\\
\pi(m,s|a)&=\frac{a}{2}\sqrt{\frac{s}{2\pi}}\; \exp(-(s/2)(m^2+a)),
\end{align}
and $q(x)=p(x|0,1)$. 
Then by using the same method as section 2.1 in \cite{wata2018} and notations
$A=\sum X_i^2$, $B=\sum X_i$, 
The free energy and the leave-one-out cross validation are derived as
respectively by
\begin{align}
F_n (a) &= - \log a +\frac{n+1}{2}\log ((A+a)(n+1)-B^2) +C_1
\\
C_n (a) 
&= 
 \frac{n+1}{2}\log ((A+a)(n+1)-B^2)\\
&
-\frac{1}{2}\sum_{i=1}^n\log((A-X_i^2+a)n-(B-X_i)^2) +C_2,
\end{align}
where either $C_1$ or $C_2$ does not depend on $a$. 
Hence the asymptotic analysis of the zero points of $F_n'(a)$ and 
$C_n'(a)$ shows that minimizers of $F_n(a)$ and $C_n(a)$ 
converge to the different values 2 and 4
as $n\rightarrow \infty$.

\vskip3mm
\noindent
{\bf Example.3 }. A simple polynomial regression is studied. 
Let $x,y\in\RR$ and $f(x)\in\RR^K$, and $a\in\RR^K$,
where $K$ is a positive integer. 
Using a function 
\[
f(x)=(1,x,x^2,...,x^{K-1})^T,
\]
a pair of a statistical model  and a prior distribution is defined by
\begin{align}
p_k(y|x,a,s)&=
\left(
\frac{s}{2\pi}
\right)^{1/2}
\exp\left(
-\frac{s}{2}(y-a\cdot f(x))^2
\right),
\\
\pi(a,s|b,c,d)&\propto 
s^{b-1}\exp\left(-cs-\frac{ds}{2}\|a\|^2\right). 
\end{align}
Note that the prior distribution is proper if and only if
the hyperparameter $(b,c,d)$ satisfies $b>K/2$, $c>0$, $d>0$. 
Let $q(x)q(y|x)$ be a data-generating distribution where $q(x)$ is
a standard normal distribution and $q(y|x)=p(y|x,a_0,s_0)$ with 
$a_0=(1,-1/5,1/30)$ and $s_0=25$. 
An experiment with $K=3$ and $n=20$ was conducted. 
The hyperparameter was set as $(b,c,d)=(b,0.01,0.01)$,
where $b=-3,-2.5,-2,...,6.5$ are candidate hyperparameters. 
The optimal hyperparameter among them that minimizes $\EE[G_n(\pi)]$ is $b=0.5$, 
which means the optimal one about $b$ makes the prior distribution improper. 
Two hundred independent trials were conducted. 
The optimal hyperparameters for leave-one-out cross validation,
widely applicable information criterion, 
adjusted cross validation, 
hold-out cross validation, and the free energy 
were chosen and the generalization errors for the
chosen hyperparameters were recorded. 
Table \ref{table:333} shows averages, standard deviation, and root
square error of the optimized hyperparameters. Also the average and
standard deviation of the corresponding 
generalization errors are shown.

The experimental results show that  the leave-one-out cross validation (LOO)
and the widely applicable information criterion (WAIC) estimated  
the optimal hyperparameter better than the other other methods and that 
the standard deviation of the chosen hyperparameters by WAIC was smaller than
LOO. It would be helpful that minimization of the free energy is different 
from that of the generalization loss.

\begin{table}[tb]
\begin{center}
\begin{tabular}{|c|c|c|c|c|c|}
\hline
\multicolumn{1}{|c|}{} &
\multicolumn{3}{|c|}{Optimized Hyper} 
& \multicolumn{2}{|c|}{Gen. Err.}\\
\cline{2-6}
 & Mean & Std & RSE & Mean & Std
\\
\hline
  LOO. E. &  0.75 &  0.81 &   0.85&  0.117 &  0.082
\\
\hline
WAIC. E. &  0.92 &  0.70&  0.81&  0.116 &  0.082
\\
\hline
AC. E.  &  1.05 &  1.10 &  1.22&  0.119&   0.087
\\
\hline
HO. E.  &  1.96 &  2.65 &  3.02 &  0.143 &  0.115
\\
\hline
Free E. &   2.23 &  0.25&   1.75 &  0.125 &   0.095
\\
\hline
\end{tabular}
\end{center}
\caption{Hyperparameter Optimization\newline
LOO.E.: Leave-one-out error,\newline
WAIC.E.: Widely applicable information criterion error,\newline
AC.E.: Adjusted cross validation error, \newline
HO.E.: Hold-out cross validation error,\newline
Free.E.: Free energy 
}

\label{table:333}
\end{table}

\section{Discussion}\label{section:discuss}

In this section, we discuss three points in Bayesian statistics in this paper.

\subsection{Uncertainty and Information Source in this Paper} 
\label{subsection:meaning}

In general, it seems that 
there is no mathematical definitions of 
``uncertainty'' and ``information source''. In this subsection, 
the meaning of these words in this paper is explained. 

In general, a sample $x^n$ is given from the real world, however, 
$x^n$ may or may not be a realization of some random variable in a 
real world. 
The terminology ``uncertainty'' in this paper is used for 
the case when a person does not know whether $x^n$ is 
a realization of some random variable or not. 

We study a case when a person who makes 
a special statistical model $p(x|\theta)$ and a special prior distribution $\pi(\theta)$ 
 is aware they are fictional candidates,
and wants to evaluate them 
from a more general viewpoint.  It is a mathematical fact 
that the general pair $(q(x),Q(q))$ contains 
a special pair $(p(x|\theta),\pi(\theta))$ as a specific one, hence
if a special pair exists, then the general pair also exists. 
The terminology ``unknown information source'' in this paper 
is used for the general pair $(q(x),Q(q))$ represented by a probability 
distribution. 

We can understand that both pairs 
are prepared in a person's mathematical analysis
 which may be different form the real world. 
If the unknown general pair cannot be assumed to exist 
in the real world, then the evaluation by cross validation and information 
criteria shows the adequateness of a special pair in a person's mathematical 
analysis. 
If the unknown general pair can be assumed to exist in the real world by some 
scientific viewpoint, then the evaluation by them has a scientifically objective 
meaning. 

Note that the same mathematical theory holds even if the unknown 
general pair exists in a real world or not, and that no concrete information  
about the unknown information source is required in the use of  the cross validation 
and information criteria. Hence, if a person cannot believe in a special pair, 
the evaluation by them is useful in both cases.

\subsection{Real Log Canonical Threshold}\label{subsection:RLCT}

In this subsection, we summarize researches about the real log
canonical threshold (RLCT). 

If $q(x)$ is regular for $p(x|\theta)$, then
$\lambda=d/2$, $m=1$, and $R_2=\mbox{tr}(IJ^{-1})/2$ in eq.(\ref{eq:Gn}), 
where $d$ is the dimension of the parameter space, 
$I$ and $J$ are defined in subsection \ref{subsection:3.3}.
If $q(x)$ is regular for and realizable by  $p(x|\theta)$, then 
$2 R_1$ is a random variable whose probability distribution converges to 
 $\chi^2$ distribution with $d$ degrees of freedom, and $R_2$ converges to $d/2$. 

If a statistical model contains hierarchical structure or hidden variable, then
it is in general singular. Such a statistical model is often prepared 
sufficiently large so that it can approximate an unknown information source,
hence RLCTs in overparametrized cases are important in model evaluation. 

The concrete values of RLCTs of singular statistical models have been 
clarified, for example, reduced rank regressions
\cite{Aoyagi2005},  neural networks \cite{Aoyagi2012,wata2001}, 
normal mixtures \cite{Yama2003}, Poisson mixtures\cite{Sato}, 
Latent Dirichlet allocations \cite{Hayashi2021}, and 
multinomial mixtures \cite{Takumi2022}. 
It is also clarified that, based on RLCT, the exchange probability of
exchange MCMC methods can be optimally designed \cite{Nagata2008}. 
The mean field approximation of the posterior distribution is called the variational Bayesian inference. 
Both the generalization loss and the free energy of the variational Bayes 
approximation are 
clarified \cite{Kazuho,Nakajima}, which are different from the exact values 
even asymptotically.


\subsection{Comparison of Cross Validation and Information Criterion}
\label{subsection:LOOdiscuss}

In this subsection, we compare cross validation and information criterion 
from theoretical and computational points of view. 

First, we compare cross validation and information criterion in not i.i.d. cases. 
If a sample is i.i.d. or exchangeable, then leave-one-out cross validation (LOO)
and WAIC are asymptotically equivalent, if otherwise not. 
For example, in a regression problem where the conditional probability 
density $\{q(Y_i|x_i)\}$ is estimated, 
 $\{x_i\}$ may be independent or not. 
If a probability distribution of $\{Y_i\}$ for fixed $\{x_i\}$ is given by
\[
\prod_{i=1}^n q(y_i|x_i),
\]
where $\{x_i\}$ is not a random variable, 
then we define the generalization loss by 
\[
G_n^{(1)}=-\frac{1}{n}\sum_{i=1}^n \int dy\;  q(y|x_i)\;\log p(y|x^n,Y^n).
\]
If a probability distribution of $\{(X_i,Y_i)\}$ is given by
\[
\prod_{i=1}^n q(x_i,y_i). 
\] 
then $\{(X_i,Y_i)\}$ are independent. We define
the generalization loss by 
\[
G_n^{(2)}=-\int dx \int dy \;q(x,y) \;\log p(y|X^n,Y^n) .
\]
 LOO and WAIC are asymptotically 
equivalent as the estimator of $G_n^{(2)}$, if $\{(X_i,Y_i)\}$ are
independent. If $\{Y_i\}$ is independent but $\{x_i\}$ is fixed, 
WAIC can be used as an estimator of $G_n^{(1)}$, whereas LOO not. 
This difference becomes large if a leverage sample point is 
contained or the dimension of inputs $\{X_i\}$ is high
\cite{watanabe2021a}. Hence, for the cases when $\{Y_i\}$ are independent, 
then the adjusted cross validation eq.(\ref{eq:An}) should be defined by the 
replacement of $C_{n_1}$ by $W_{n_1}$. 
For general dependent cases such as time sequence analysis, 
the evaluation methods of the statistical models and prior distributions are 
not yet fully 
constructed, however, several measures which contain a specific statistical
model, for example, the leave-future-out cross validation, were 
 proposed \cite{Burkner2020}.

Second,  we study computational problems of cross validation and information
criterion. 
In Bayesian inference, calculation of the leave-one-out cross validation 
needs heavy computational
costs because the posterior distributions for $X^n\setminus X_i$ is necessary
for all $i=1,2,...,n$. The importance sampling 
cross validation 
\begin{align}\label{eq:ISCV}
{\rm ISCV}=\frac{1}{n}\sum_{i=1}^n \EE_\theta[\log(1/p(X_i|\theta))]
\end{align}
can be used for numerical approximation.
If a leverage sample point is contained a sample, 
the posterior average or variance may be infinite cite{Epifani,Perugia}, and 
a new estimation method has been proposed which are useful in many 
practical problems  \cite{Vehtari1,VehtariPSIS,Vehtari2}. 

\section{Conclusion}\label{section:conclusion}

We introduced a place of 
mathematical theory of Bayesian statistics when uncertainty is unknown. 
In almost all cases, a person who makes a pair of a statistical model and
a prior distribution is aware that it is a  fictional  candidate.
Therefore, a person needs statistical evaluation measures which can be employed
even if an uncertainty is unrealizable or singular for a statistical model. 
Mathematical theory of 
a cross validation, an information criterion, and a marginal likelihood 
will be one of the most important foundation for Bayesian statistics 
for unknown information source.


\begin{thebibliography}{99}

\bibitem{Akaike1974}
Akaike, H. A new look at the statistical model identification, 
 IEEE Transactions on Automatic Control. Vol.19, No.6,  1974.

\bibitem{Akaike1980}
Akaike, H.
On the transition of the paradigm of statistical inference.
The proceedings of the Institute of Statistical Mathematics,
Vol.27,	 pp.5-12. 1980. 

\bibitem{Akaike1980a}
Akaike, H. Likelihood and Bayes procedure. Bayesian Statistics. pp. 143-166, 1980.

\bibitem{Aoyagi2005}
Aoyagi, M., Watanabe, S. Stochastic complexities of reduced rank regression in Bayesian estimation. Neural Networks, Vol.18, pp.924-933, 2005.

\bibitem{Aoyagi2012}
Aoyagi, M., Nagata, K. Learning coefficient of generalization error in Bayesian estimation and Vandermonde matrix type singularity. Neural Computation, vol. 24, No. 6, pp.1569 -1610, 2012.

\bibitem{Atiyah}
 Atiyah. M.F. Resolution of singularities and division of distributions
communications on pure and applied mathematics. Vol.23, no.2. pp.145-150. 1970.

\bibitem{Binmore2017}
Binmore, K. On the foundations of decision theory. Homo Oeconomicus, 
Vol.34, pp.259-273, 2017.


\bibitem{Box1976}
Box, G. E. P. Science and statistics. Journal of the American Statistical Association. Vol.71, pp.791-799, 1976. 


\bibitem{Burkner2020}
B\"{u}rkner. P.-C., Gabry, I.-J., Vehtari, A. 
Approximate leave-future-out cross-validation for Bayesian time series models.
Journal of Statistical Computation and Simulation. 
Vol.90,  pp.2499-2523, 2020. 

\bibitem{Drton}
Drton, M., Plummer, M. A Bayesian information criterion for singular models. J. R. Statist. Soc. B., Vol.56, pp.1-38, 2017.

\bibitem{Efron}
Efron,B., Hastie, T. Computer age statistical inference : 
algorithms, evidence, and data Science. 
Cambridge University Press, 2016.


\bibitem{Epifani}
Epifani, I., MacEchern, S. N., Peruggia, M. Case-deletion 
importance sampling estimators: central limit 
theorems and related results. Electric Journal of 
Statistics, Vol.2, pp.774-806, 2008.

\bibitem{Gelfand1992}
Gelfand, A. E., Dey, D. K., Chang, H. 
Model determination using predictive distributions with
  implementation via sampling-based method. 
Technical Report, Department of statistics,
Stanford University, Vol.462, pp. 147-167, 1992.

\bibitem{GelmanBDA} Gelman, A.,
Carlin, J. B.,
Stern H.S.,
Dunson, D.B.,
Vehtari, A.,
Rubin D.B.
Bayesian data analysis III. CRC Press. 2013. 

\bibitem{Gelman2013}
Gelman, A., Shalizi, C. S. 
Philosophy and the practice of Bayesian statistics. 
British Journal of Mathematical and Statistical Psychology. Vol.66, pp.8-38, 2013.

\bibitem{Gelman3}
Gelman, A., Hwang, J., Vehtari, A.
 Understanding predictive information criteria for Bayesian models.
Statistics and Computing, Vol.24, pp.997--1016, 2014.


  
\bibitem{Hayashi2021}
Hayashi, N. 
The exact asymptotic form of Bayesian generalization error in latent Dirichlet allocation, Neural Networks, Vol.137, pp.127-137, 2021. 

\bibitem{Hironaka}
Hironaka, H.
Resolution of singularities of an algebraic variety over a field of characteristic zero. I,II.  Ann. of Math., Vol.79, pp.109-326, 1964. 

\bibitem{McElreath2020}
McElreath, S.
Statistical Rethinking: A Bayesian course with examples in R and STAN. 2nd edition. CRC Press, 2020. 

\bibitem{Nagata2008}
Nagata, K. Watanabe, S. 
Asymptotic behavior of exchange ratio in exchange Monte Carlo method. 
 Neural Networks, Vol. 21, pp. 980-988, 2008.

\bibitem{Nagayasu}
Nagayasu, S. Watanabe, S. 
Asymptotic behavior of free energy when optimal probability distribution is not unique. 
Neurocomputing, Vol.500,pp.528-536, 2022. 

\bibitem{Nakajima}
Nakajima,S. Watanake,K. Sugiyama,M. Variational Bayesian Learning Theory, Cambridge University Press, 
2019. 

\bibitem{Peruggia}
 Peruggia, M.
 On the variability of case-detection importance sampling weights in
  the Bayesian linear model.
Journal of American Statistical Association, Vol.92, 
pp.199-207, 1997.


\bibitem{Sato}
Sato, K., Watanabe, S. 
Bayesian generalization error of Poisson mixture and simplex Vandermonde 
matrix type singularity. arXiv:1912.13289, 2019. 

\bibitem{Schwarz1978}
Schwarz, G. 
Estimating the dimension of a model.  Annals of Statistics, Vol. 6, pp.461-464. 1978. 

\bibitem{Spiegel2002}
Spiegelhalter, D.J., Best, N. G.,
Carlin, B. P., Linde, A. Bayesian measures of model complexity and
fit. Journal of Royal Statistical Society, Series B, Vol.64, pp.583-639, 2002. 


\bibitem{Vehtari1}
Vehtari, A., Lampinen, J.
Bayesian model assessment and comparison using cross-validation
  predictive densities.
Neural Computation, Vol.14, pp.2439--2468, 2002.



\bibitem{VehtariPSIS}
Vehtari,A., Simpson, D., Gelman, A., Yao, Y., Gabry, J.,
Pareto smoothed importance sampling. 
arXiv:1507.02646v8, 2015. 



\bibitem{Vehtari2}
Vehtari, A., Gelman, A., Gabry, J. 
Practical Bayesian model evaluation using leave-one-out cross-validation and WAIC.
Statistics and computing. Vol. 27, pp.1413-1432, 2017. 

\bibitem{Smith}
Smith, A. F. M. 
Present position and potential developments: 
Some personal views: Bayesian Statistics. 
Journal of the Royal Statistical Society. Series A (General) , Vol. 147, 
pp. 245-259, 1984.

\bibitem{Kazuho}
Watanabe,K.   Watanabe,S.
Stochastic complexities of Gaussian mixtures in variational Bayesian approximation.
Journal of Machine Learning Research
Vol.7, pp.625-644, 2006. 

\bibitem{wata2001}
Watanabe, S. Algebraic geometrical methods for hierarchical learning machines.
Neural Networks, Vol.14, pp.1049-1060, 2001. 

\bibitem{wataamari2003}
Watanabe,S. Amari, S. 
Learning coefficients of layered models when the true
distribution mismatches the singularities. 
Neural Computation, Vol.15, pp. 1013-1033, 2003. 

\bibitem{Watanabe2009} 
Watanabe, S. Algebraic geometry and statistical learning theory.
Cambridge University Press. 2009. 

\bibitem{jmlr1}Watanabe, S. Asymptotic equivalence of Bayes cross validation and widely
applicable information criterion in singular learning theory. Journal of Machine Learning Research. Vol.11, pp.3571-3594, 2010.

\bibitem{renormalizable}
Watanabe, S. Asymptotic learning curve and renormalizable condition in statistical learning theory, Journal of Physics Conference Series, Vol. 233, No. 1, 2010.

\bibitem{jmlr2}Watanabe, S. 
A widely applicable Bayesian information criterion. Journal of Machine Learning Research. Vol. 14, pp.867-897, 2013. 

\bibitem{wata2018} 
Watanabe, S. Mathematical theory of Bayesian statistics. 
CRC Press, 2018. 

\bibitem{watanabe2018}
Watanabe, S. Higher order equivalence of Bayes cross validation and 
WAIC. Proceedings in Mathematics and Statistics, Springer, 
Information Geometry and Its Applications, pp.47-73, 2018. 

\bibitem{watanabe2021a}
Watanabe, S. WAIC and WBIC for mixture models. Behaviormetrika,
doi.org$/$10.1007$/$s41237-021-00133-z, 2021. 

\bibitem{watanabe2021b}
Watanabe,S. 
Information criteria and cross validation for Bayesian inference in regular and singular cases. 
Japanese Journal of Statistics and Data Science volume,Vol 4, pp.1-19, 2021.

\bibitem{Watanabe2022}
Watanabe, S. Mathematical theory of Bayesian statistics where all models are wrong. Advancements in Bayesian Methods and Implementations, Handbook of statistics, Vol.47, Elsevier, 2022.

\bibitem{Takumi2022} Watanabe,T., Watanabe,S. 
Asymptotic behavior of Bayesian generalization error in multinomial mixtures.
arXiv:2203.06884. 

\bibitem{Yama2013}
Yamazaki, K., Kaji, D. 
Comparing two Bayes methods based on the free energy functions in Bernoulli mixtures.
Neural Networks. Vol.44, pp.36-43, 2013. 

\bibitem{Yama2003}
Yamazaki, K., Watanabe, S. Singularities in mixture models and upper bounds of stochastic complexity. Neural Networks. 
Vol.16, pp.1029-1038, 2003.



\end{thebibliography}
\end{document}